\documentclass[conference]{IEEEtran}

\ifCLASSINFOpdf
   \usepackage[pdftex]{graphicx}
\else
\fi

\usepackage{cite}
\usepackage{amsmath}
\usepackage{algorithmic}
\usepackage{graphicx,subcaption}
\usepackage{color}
\newcommand\T{\rule{0pt}{2.6ex}}       % Top strut
\newcommand\B{\rule[-1.2ex]{0pt}{0pt}} % Bottom strut
\newcommand{\Ni}{({\em i})~}
\newcommand{\Nii}{({\em ii})~}

\begin{document}
\title{Robust Automated Human Activity Recognition\\and its Application to Sleep Research}

\author{
\IEEEauthorblockN{Aarti Sathyanarayana}
\IEEEauthorblockA{Qatar Computing Research Institute\\
Hamad Bin Khalifa University\\
Doha, Qatar}\\   %<------ Line breaks in the current column
\IEEEauthorblockN{Jaideep Srivastava}
\IEEEauthorblockA{Qatar Computing Research Institute\\
Hamad Bin Khalifa University\\
Doha, Qatar}
\and
\IEEEauthorblockN{Ferda Ofli}
\IEEEauthorblockA{Qatar Computing Research Institute\\
Hamad Bin Khalifa University\\
Doha, Qatar}\\ %[0.9cm]  <------- Extra vertical space
\IEEEauthorblockN{Ahmed Elmagarmid}
\IEEEauthorblockA{Qatar Computing Research Institute\\
Hamad Bin Khalifa University\\
Doha, Qatar}
\and
\IEEEauthorblockN{Luis Fernandez-Luque}
\IEEEauthorblockA{Qatar Computing Research Institute\\
Hamad Bin Khalifa University\\
Doha, Qatar}\\                 %<-----------
\IEEEauthorblockN{Teresa Arora\\
and Shahrad Taheri}
\IEEEauthorblockA{Weil Cornell Medical College - Qatar\\
Qatar Foundation\\
Doha, Qatar}
}

\maketitle
\thispagestyle{plain}
\pagestyle{plain}

\begin{abstract}
Human Activity Recognition (HAR) is a powerful tool for understanding human behaviour. Applying HAR to wearable sensors can (1) provide new insights by enriching the feature set in health studies, and (2) enhance the personalisation and effectiveness of health, wellness, and fitness applications. Wearable devices provide an unobtrusive platform for user monitoring, and due to their increasing market penetration, feel intrinsic to the wearer. The integration of these devices in daily life provide a unique opportunity for understanding human health and wellbeing. This is referred to as the ``quantified self" movement.

The analyses of complex health behaviours such as sleep, traditionally require a time-consuming manual interpretation by experts. This manual work is necessary due to the erratic periodicity and persistent noisiness of human behaviour. In this paper, we present a robust automated human activity recognition algorithm, which we call RAHAR. We test our algorithm in the application area of sleep research by providing a novel framework for evaluating sleep quality and examining the correlation between the aforementioned and an individual's physical activity. Our results improve the state-of-the-art procedure in sleep research by 15\% for area under ROC and by 30\% for F1 score on average. However, application of RAHAR is not limited to sleep analysis and can be used for understanding other health problems such as obesity, diabetes, and cardiac diseases.

%Human Activity Recognition (HAR) on wearable sensors is a powerful tool for (1) medical diagnostic screening processes that currently are limited to manual and brief in-clinic evaluations, and (2) user empowerment applications for an improved quality of life. Wearable devices provide an unobtrusive platform for user monitoring, and due to their increasing market penetration feel intrinsic to the wearer. This daily integration into a person's life is crucial for increasing the understanding of overall human health and wellbeing. In this paper, we present a robust automated human activity recognition algorithm (RAHAR) for multi-modal phenomena, such as activity data. We test our model in the application area of sleep science by providing a novel framework for evaluating sleep quality and examining the correlation between the aforementioned and an individual's daily physical activity. Our results show a 30\% improvement over the current state-of-the-art procedures. More generally, using our HAR approach with a wearable device empowers users to self-monitor their own sleep patterns, and reform their activity habits for optimised sleep and an improved quality of life. ADD SOMETHING FROM DISCUSSION OF BEST RESULTS
\end{abstract}

\IEEEpeerreviewmaketitle

\section{Introduction}
Human Activity Recognition (HAR) is the understanding of human behaviour from data captured by pervasive sensors, such as cameras or wearable devices. It is a powerful tool in medical application areas, where consistent and continuous patient monitoring can be insightful. Wearable devices provide an unobtrusive platform for such monitoring, and due to their increasing market penetration, feel intrinsic to the user. This daily integration into a user's life is crucial for increasing the understanding of overall human health and wellbeing. This is referred to as the ``quantified self" movement. 

%Wearables, such as actigraph accelerometers, present a continuous time series of a person's daily physical exertion and rest. This extended monitoring cannot be replicated in a clinical setting, and can generate large amounts of data that provide new insights for medical screening processes that currently require surveillance in a lab setting and elaborate at-home equipment. Moreover, this data is not collected 

Wearables, such as actigraph accelerometers, generate a continuous time series of a person's daily physical exertion and rest. This ubiquitous monitoring presents substantial amounts of data, which can \Ni provide new insights by enriching the feature set in health studies, and \Nii enhance the personalisation and effectiveness of health, wellness, and fitness applications. By decomposing an accelerometer's time series into distinctive activity modes or actions, a  comprehensive understanding of an individual's daily physical activity can be inferred. The advantages of longitudinal data are however complemented by the potential of noise in data collection from an uncontrolled environment. Therefore, the data sensitivity calls for robust automated evaluation procedures.

%The implicit obstacle to HAR and segmentation is the lack of regular periodicity in human behaviour. 

In this paper, we present a robust automated human activity recognition (RAHAR) algorithm. We test our algorithm in the application area of sleep science by providing a novel framework for evaluating sleep quality and examining the correlation between the aforementioned and an individual's physical activity. Even though we evaluate the performance of the proposed HAR algorithm on sleep analysis, RAHAR can be employed in other research areas such as obesity, diabetes, and cardiac diseases.

\section{Related Work}

Human activity recognition (HAR) has been an active research area in computer vision and machine learning for many years. A variety of approaches have been investigated to accomplish HAR ranging from analysis of still images and videos to motion capture and inertial sensor data.
 
Video has been the most widely studied data source in HAR literature. Hence, there exists a wealth of papers in this particular domain. The most recent literature on HAR from videos include trajectory-based descriptors~\cite{ HWang:IJCV15, BFernando:TPAMI16, IAtmosukarto:WACV15}, spatio-temporal feature representations~\cite{ SMa:CVPR15, ZZhou:TMM15, DTran:ICCV15}, feature encoding~\cite{ VKantorov:CVPR14, XPeng:ECCV14, HKuehne:WACV16}, and deep learning~\cite{ JDonahue:CVPR15, LWang:CVPR15, LSun:ICCV15}. Reviewing the extensive list of video-based HAR studies, however, goes beyond the scope of this study and we refer the reader to~\cite{ SKe:Computers13, PBorges:TCSVT13} for a collection of more comprehensive surveys on the topic. 

Unlike HAR from video, existing approaches for HAR from still images are somewhat limited, and range from histogram-based representations~\cite{NIkizler:ICPR08, CThurau:CVPR08} and color descriptors~\cite{FKhan:IJCV13} to pose-, appearance- and parts-based representations~\cite{WYang:CVPR10, SMaji:CVPR11, BYao:ICCV11, GSharma:TPAMI16}. Guo and Lai recently provided a comprehensive survey of the studies on still image-based HAR in~\cite{GGuo:PatRec14}.

Several techniques have been proposed, on the other hand, for HAR from 3D data, encompassing representations based on bag-of-words~\cite{ WLi:CVPRW10, LXia:CVPRW12}, eigen-joints~\cite{ XYang:CVPRW12}, sequence of most informative joints~\cite{ FOfli:JVCI14}, linear dynamical systems~\cite{ RChaudhry:CVPRW13}, actionlets~\cite{ JWang:CVPR12}, Lie algebra embedding~\cite{ RVemulapalli:CVPR14}, covariance descriptors~\cite{ MHussein:IJCAI13}, hidden Markov models~\cite{ FLv:ECCV06}, subspace view-invariant metrics~\cite{ YSheikh:ICCV05} and occupancy patterns~\cite{ JWang:ECCV12, AVieira:CIARP12}. Aggarwal and Xia presented a recent survey summarizing state-of-the-art techniques in HAR from 3D data~\cite{JAggarwal:PatRec14}.

Unlike vision-based HAR systems, sensor-based HAR technologies commonly deal with time series of state changes and/or various parameter values collected from a wide range of sensors such as contact sensors, accelerometers, audio and motion detectors, etc. Chen et al.~\cite{LChen:TCMCC12} and Bulling et al.~\cite{ABulling:CSUR14} present comprehensive reviews of sensor-based activity recognition literature. The most recent work in this domain includes knowledge-based inference~\cite{ACalzada:EMBC14, DBiswas:HUMOV15}, ensemble methods~\cite{ATripathi:EAIS15, CCatal:ASOC15}, data-driven approaches~\cite{RAkhavian:WSC15, LLiu:KNOSYS15}, and ontology-based techniques~\cite{GOkeyo:PMCJ14}.

All of the aforementioned studies investigate recognition/classification of fully observed action or activity, e.g., jumping, walking, running, drinking, etc. (i.e., activities of daily living), using well-curated datasets. However, thanks to the ``quantified self" movement, myriad of consumer-grade wearable devices have become available for individuals who have started monitoring their physical activity on a continuous basis, generating tremendous amount of data. Therefore, there is an urgent need for automatic analysis of data coming from fitness trackers to assess the physical activity levels and patterns of individuals for the ultimate goal of quantifying their overall wellbeing. This task requires understanding of longitudinal, noisy physical activity data at a rather higher (coarser) level than specific action/activity recognition level. Main challenges as well as opportunities of HAR from personalized data and lifelogs have been discussed in several dimensions in~\cite{BDobkin:CurrOpinNeurol13, OLara:SURV13, JBort-Roig:Sports14, MRehman:Sensors15, SMukhopadhyay:JSEN15}.

There has been a number of initiatives to overcome the challenge of collecting annotated personalized data to further research on HAR from continuous measurement of real-world physical activities~\cite{MZhou:ICMR13, CMeurisch:UbiComp15}. Even though such systems exhibit a crucial attempt in furthering research in mining personalized data, they have limited practical importance as they rely on manual annotation of the acquired data. There has also been recent attempts to automatically recognize human activities from continuous personalized data~\cite{JHamm:MobiCASE13, CDobbins:CIT15, MUddin:WearSys15, OBanos:EMBC15}. However, most of these studies are designed to recognize only a predefined set of activities, and hence, not comprehensive and robust enough to quantify the physical activity levels for the overall assessment of individuals' wellbeing.

\section{Background}
Sleep pattern evaluation is a paragon of cumbersome testing and requires extensive manual evaluation and interpretation by clinical experts. Unhealthy sleep habits can impede physical, mental and emotional wellbeing, and lead to exacerbated health consequences~\cite{strine2005associations}. Since patient referral to sleep specialists is often based on self-reported abnormalities, exacerbation often precedes diagnosis. 

Clinical diagnosis of complex sleep disorders involves a variety of tests, including an overnight lab stay with oxygen and brain wave monitoring (polysomnography and electroencephalogram, respectively), and a daily sleep history log with a subjective questionnaire. The daily sleep logs and questionnaires are often found to be unreliable and inconsistent with actual observed activity. This is especially true in adolescents~\cite{arora2013investigation}. The overnight stay allows specialists to manually monitor the patient's sleep period. This requires the active involvement of a clinical sleep specialist. Furthermore, the monitoring is only for one night and in a clinical setting, rather than the patient's own home. Using wearable devices provides both a context-aware and longitudinal monitoring. 

The inconvenience and inaccuracy of daily logs, coupled with the invasiveness of an overnight lab stay, substantiate the need and adoption of wearable devices for first pass diagnostic screening. More generally, using our HAR approach with a wearable device empowers users to self-monitor their sleep patterns, and reform their activity habits for optimised sleep and an improved quality of life.

\section{Preliminaries}
In this section we present a description of the dataset and the context-aware definitions used for our application area.

\subsection{Data}
%The data for this analysis was collected as part of a clinical trial held by XXXX. Two high schools were selected for cohort development. Volunteers with parental consent were provided an actigraph accelerometer, ActiGraph GT3X+\footnote{http://actigraphcorp.com/support/activity-monitors/gt3xplus/}, to wear on their non-dominant wrist, continuously throughout the trial (i.e. even when sleeping). 
Data was collected as part of a research study to examine the impact of sleep on health and performance in adolescents by Weil COrnell Medical COllege - Qatar. Two international high schools were selected for cohort development. Student volunteers were provided with an actigraph accelerometer, ActiGraph GT3X+\footnote{http://actigraphcorp.com/support/activity-monitors/gt3xplus/}, to wear on their non-dominant wrist, continuously throughout the study (i.e. even when sleeping). Deidentified data collected in the study were used in the current analysis.

The ActiGraph GT3X+ is a clinical-grade wearable device that has been previously validated against clinical polysomnography~\cite{PFreedson:MedSciSports98}. The device samples the user's sleep-wake activity at 30-100 Hertz. Currently sleep experts use this device in conjunction with the accompanying software, ActiLife\footnote{http://actigraphcorp.com/products-showcase/software/actilife/}, to evaluate an individual's sleep period. We evaluate our results side-by-side with ActiLife's results. 

\subsection{Definitions}

\begin{figure*}
\caption{Sleep science definitions on an example accelerometer data extract}
\centering
   \includegraphics[scale=0.25]{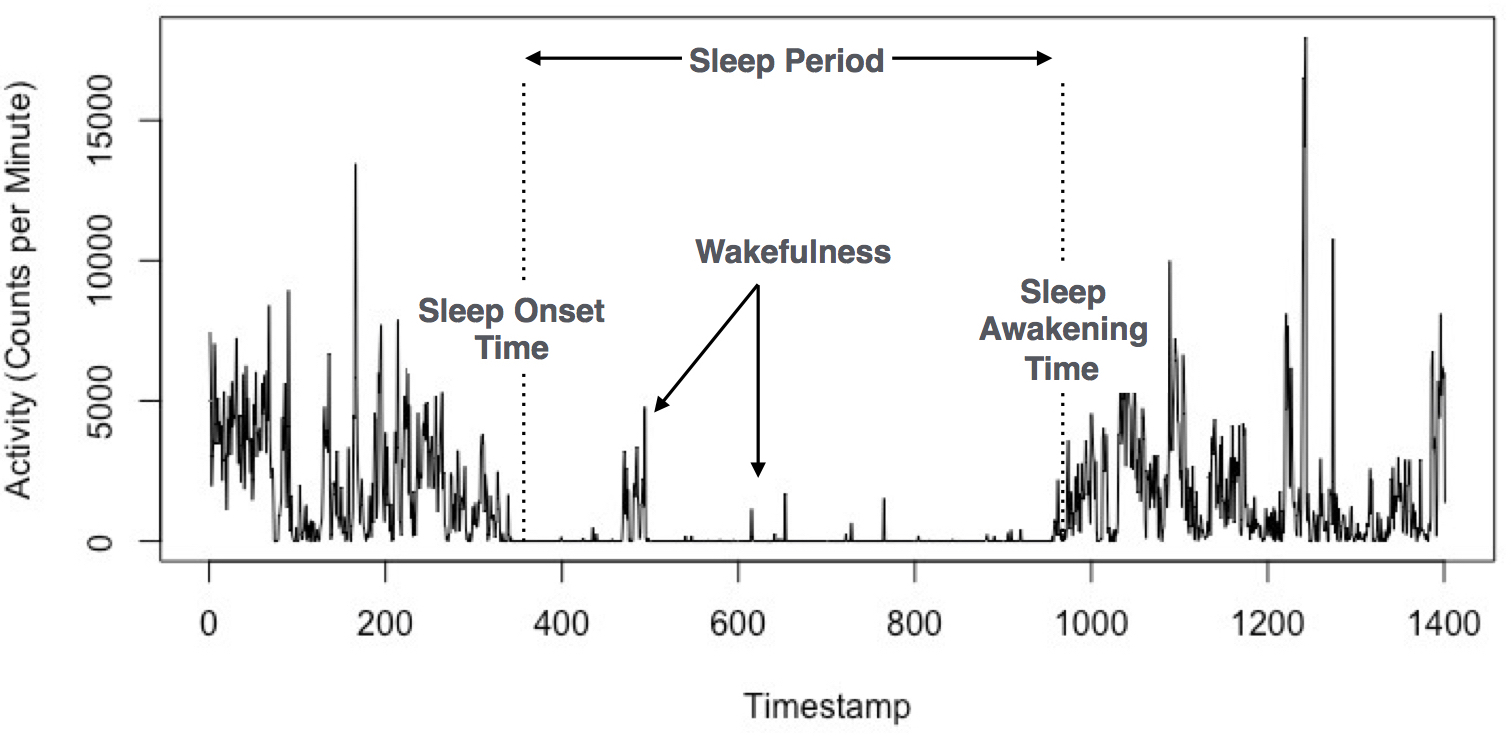}
%  \caption{Sleep science definitions on an example accelerometer data extract}
\label{fig:sleep_example}
\end{figure*}

\begin{table*}
\centering
\caption{Relevant sleep science equations~\cite{CIber:AASM07}}
\centering
\begin{tabular}{c|c}
    \hline
    Sleep Period &  $ \big[ \text{Sleep Onset Time}, \text{Sleep Awakening Time} \big] $ \T\B \\
    \hline
    Sleep Period Duration  &   $ \left \| \text{Sleep Awakening Time} - \text{Sleep Onset Time}  \right \| $ \T\B \\
    \hline
    Wake After Sleep Onset (WASO)    &  $ \sum_{\text{n}=\text{onset}}^{\text{awake}} \left \| \text{Wakefulness} \right \| $ \T\B \\
    \hline
    Latency    &  $ \big[ \text{Preceding Sedentary Time}, \text{Sleep Onset Time} \big] $ \T\B \\
    \hline
    Total Minutes in Bed    &   $ \left \| \text{Sleep Awakening Time} - \text{Preceding Sedentary Time} \right \| $ \T\B \\
    \hline
    Total Sleep Time    &   $ \left \| \text{Sleep Period Duration} - \text{WASO} - \text{Latency}  \right \| $ \T\B \\
    \hline
    Sleep Efficiency    &  $ \text{Total Sleep Time} / \text{Total Minutes in Bed}  $ \T\B \\
    \hline
\end{tabular}
\label{tab:sleep_eqs}
\end{table*}

To apply our methodology to the area of sleep science, it is important to note the definitions mentioned in this section. In traditional sleep study literature, a sleep period is bounded between the sleep-onset-time and sleep-awakening-time~\cite{CIber:AASM07}. Experts characterise the sleep-onset-time as the first minute after a self-reported bedtime, that is followed by 15 minutes of continuous sleep~\cite{sadeh2000sleep}. We propose a modified definition, that allows for automatic evaluation and deems sleep diaries unnecessary. As a result, we can infer the ``bedtime" of an individual in reverse, based on their sedentary activity before the onset of sleep. Epoch records that contain no triaxial movement, 0 steps taken, and an inclinometer output of not lying down, are candidate sleep records, and are further tested  for whether they are a component of the sleep period. We define sleep-onset-time as the first candidate epoch record in a series of 15 continuous candidate sleep minutes. Likewise, the sleep-awakening-time is defined as the last epoch record in a series of 15 continuous candidate sleep minutes, that is followed by 30 continuous non-candidate sleep minutes, (i.e. 30 minutes of active awake time). The sleep period duration can be computed as the time passed between sleep onset and sleep awakening.

Within the sleep period, there are periods of unrest or wakefulness. For example, when a user re-adjusts positions, or uses the bathroom. If the duration of movement exceeds 5 consecutive minutes of activity, it is marked as a time of ``wakefulness." The total sum of all moments of wakefulness is referred to as wake-after-sleep-onset, also known as WASO. 

Immediately preceding the start of the sleep onset, is the time-in-bed, which quantifies the sedentary time an individual spends before they have fallen asleep. This sedentary time can be observed in the actigraph accelerometer data. The time that the preceding sedentary activity begins until the time of the sleep onset is called the sleep latency. 

From the aforementioned values, total sleep time and an overall sleep efficiency score can be deduced. Total sleep time covers the defined sleep period, less the wake after sleep onset time and less the latency. Lastly, sleep efficiency is the ratio of total sleep time to total minutes in bed. All of the above definitions are summarised in Table~\ref{tab:sleep_eqs}, and visualised in Fig.~\ref{fig:sleep_example}. In this study, we use sleep efficiency as the metric to measure sleep quality~\cite{JCacioppo:PsycSci02} among other metrics such as latency, wake after sleep onset, awakening index, total sleep time, etc.~\cite{SScholle:SleepMedicine11}.

\section{Methodology}
Our methodology for RAHAR is shown algorithmically in Fig.~\ref{algo:rahar}. We elaborate on the details of our algorithm in the sequel.
%INSERT ALGORITHM FOR RAHAR, abstract all sleep specific details
%1. create granularity to epoch
%2. troiano cut points to label each epoch
%3. change point detection
%4. statistical mode of each label over the change point interval
\begin{figure}
\begin{algorithmic}[1]
\STATE{\textbf{input:} Raw accelerometer data}
\STATE{\textbf{output:} Time-series segments with activity intensity level annotations}
\FORALL{segment (daily or otherwise)}
	\FORALL{epoch (minutes, hour, etc.)}
		\STATE{implement activity cut points}
	\ENDFOR
	\STATE{change points $\gets$ implement hierarchical divisive estimation}
	\STATE{change point intervals $\gets$ divide time series by change points}
\ENDFOR
\FORALL{change point interval}
	\STATE{activity mode $\gets$ statistical mode of cut points}
\ENDFOR
\end{algorithmic}
\caption{Algorithm for Robust Automated Human Activity Recognition (RAHAR).}
\label{algo:rahar}
\end{figure}

%\begin{algorithm}
%\% Annotate physical activity time-series data\\
%\textbf{INPUT:} Raw accelerometer data\\
%\textbf{OUTPUT:} Time-series segments with activity intesity level annotations
%\begin{algorithmic}
%%\STATE Parallel Merge Sort set $S_o$ of overlay polygons based on X co-ordinates of bounding boxes\footnotemark[1]
%\FORALL{segment (daily or otherwise)}
%%\STATE find $S_x \subset S_o$ such that $B_i$ intersects with all elements of $S_x$ over $X$ co-ordinate
%	\FORALL{epoch (minutes, hour, etc.)}
%%\IF {$B_i$ intersects $O_j$ over $Y$ co-ordinate}
%		\STATE{implement activity cut points}
%%\ENDIF
%	\ENDFOR
%	\STATE{change points $\gets$ implement hierarchical divisive estimation}
%	\STATE{change point intervals - divide time series by change points}
%\ENDFOR
%\FORALL{change point interval}
%	\STATE{activity mode $\gets$ statistical mode of cut points}
%\ENDFOR
%\end{algorithmic}
%\caption{Robust Automated Human Activity Recognition (RAHAR)}
%\label{algo:relgraph}
%\end{algorithm}

%\begin{algorithm}
%\textbf{Input:} Raw accelerometer data
%\textbf{Output:} Time series with activity label annotation
%\FORALL{segment (daily or otherwise)}
%	\FORALL{epoch (minutes, hour, etc.)}
%		\STATE{Implement activity cut points}
%	\STATE{change points <- Implement hierarchical divisive estimation}
%	\STATE{change point intervals - Divide time series by change points}
%	\FORALL{change point interval}
%		\STATE{Activity mode <- statistical mode of cut points}
%\end{algorithm}

\begin{figure*}
\centering
  \caption{Classification labelling of each change point interval during an example awake time}
  \includegraphics[width=1.0\linewidth]{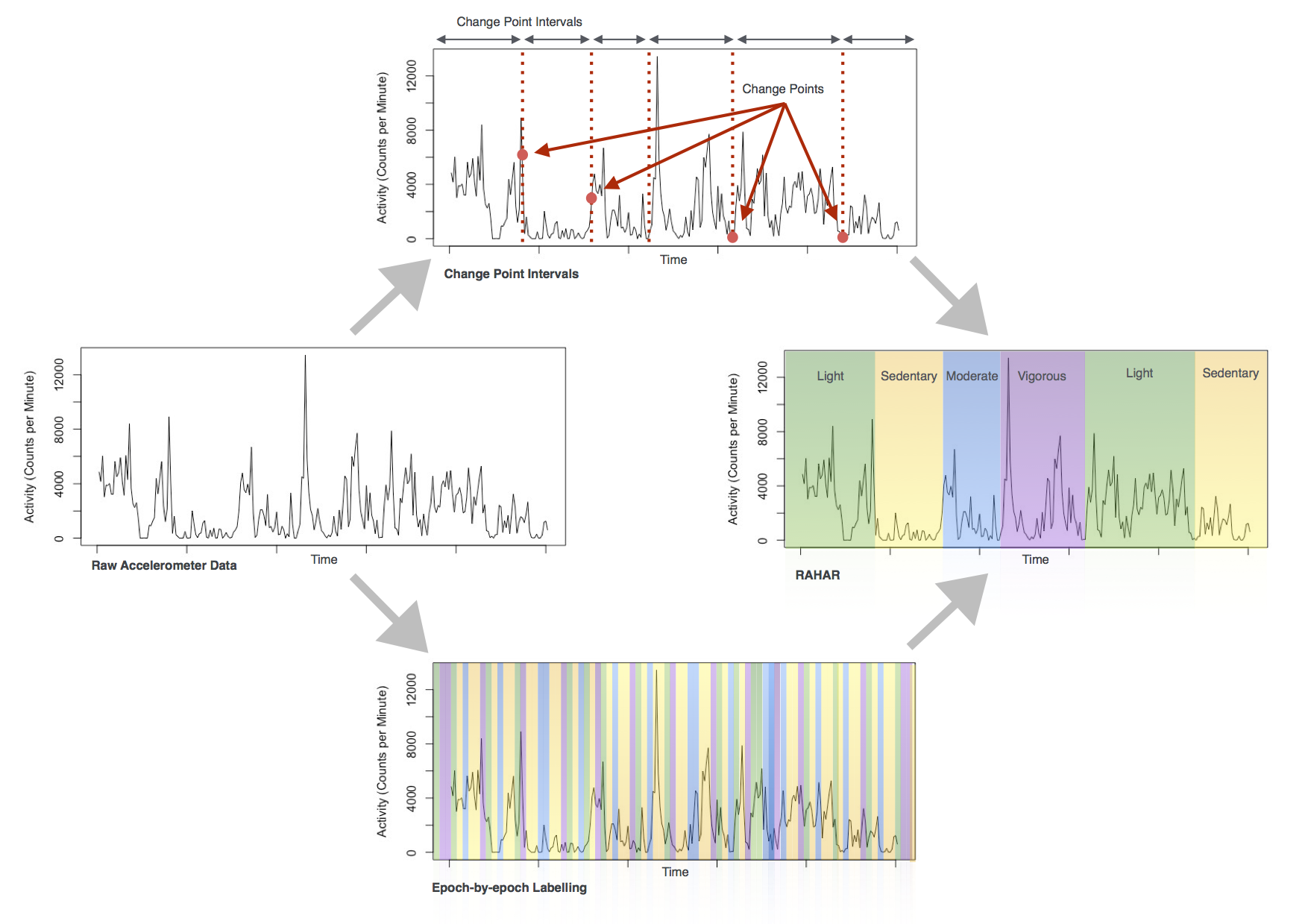}
  \label{fig:change_point_labeling}
\end{figure*}

\subsection{Pre-Processing}
The accelerometer of choice, Actigraph GT3X+, sampled each person's activity at 30-100 Hertz. The stored data included the triaxial accelerometer coordinates as well as a computed epoch step count based on the vertical axis, and post-processed inclinometer orientation.  This raw data was downloaded and aggregated to a minute-by-minute granularity. An epoch of one minute was selected in order to optimise the interpretability of the physical activity~\cite{KGabriel:IJBNPA10}, as well as for implementing the state-of-the-art cut point methodology~\cite{troiano2008physical}. In other contexts, a different granularity may be sufficient.

\subsection{Automated Annotation and Segmentation}
Due to the context of sleep disorders, sleep periods needed to be annotated within the raw ActiGraph output. Candidate sleep records, epochs with no triaxial movement, 0 steps taken, and an inclinometer output of not lying down, were identified in the time series and tested to find the sleep onset time, and sleep awakening time. The details of this terminology is elaborated in the preliminaries section. All test instances that fell within these two boundary times, were annotated as ``Sleep," and constituted the sleep period.  

Whilst analysing the data, we found that several participants had multiple sleep periods in a day, implying that they took daily naps or followed a polyphasic, or biphasic, sleep pattern. Upon closer analysis of the length and time of the sleep period, no discernible patterns were visible. Thus we opted to segment the time series by the end of a sleep period rather than the traditional approach of segmenting by day. Each sleep period was linked to its preceding activity, extending until the previous sleep period. We refer to these segments as sleep-wake segments. The result of this decision is that the activity immediately before each sleep period is used for the correlation analysis for its subsequent sleep period, rather than the total for that day. 

\begin{figure*}
    \centering
     \caption{ROC curves for sleep efficiency}
    \begin{subfigure}[b]{0.49\linewidth}        %% or \columnwidth
        \centering
        \caption{RAHAR}
        \includegraphics[width=\linewidth]{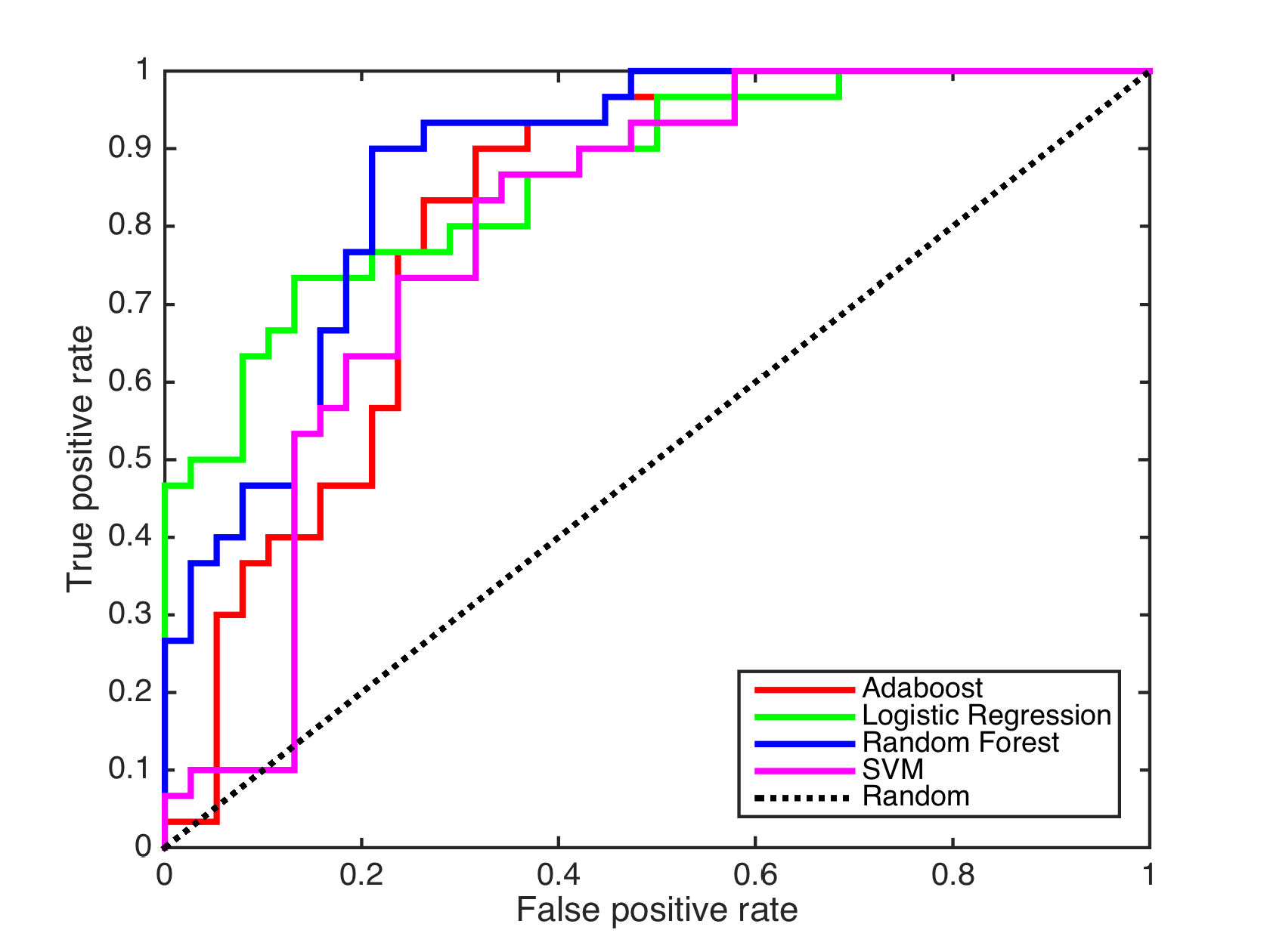}
        \label{fig:roc_RAHAR}
    \end{subfigure}
    \begin{subfigure}[b]{0.49\linewidth}        %% or \columnwidth
        \centering
        \caption{Sleep Expert + ActiLife}
        \includegraphics[width=\linewidth]{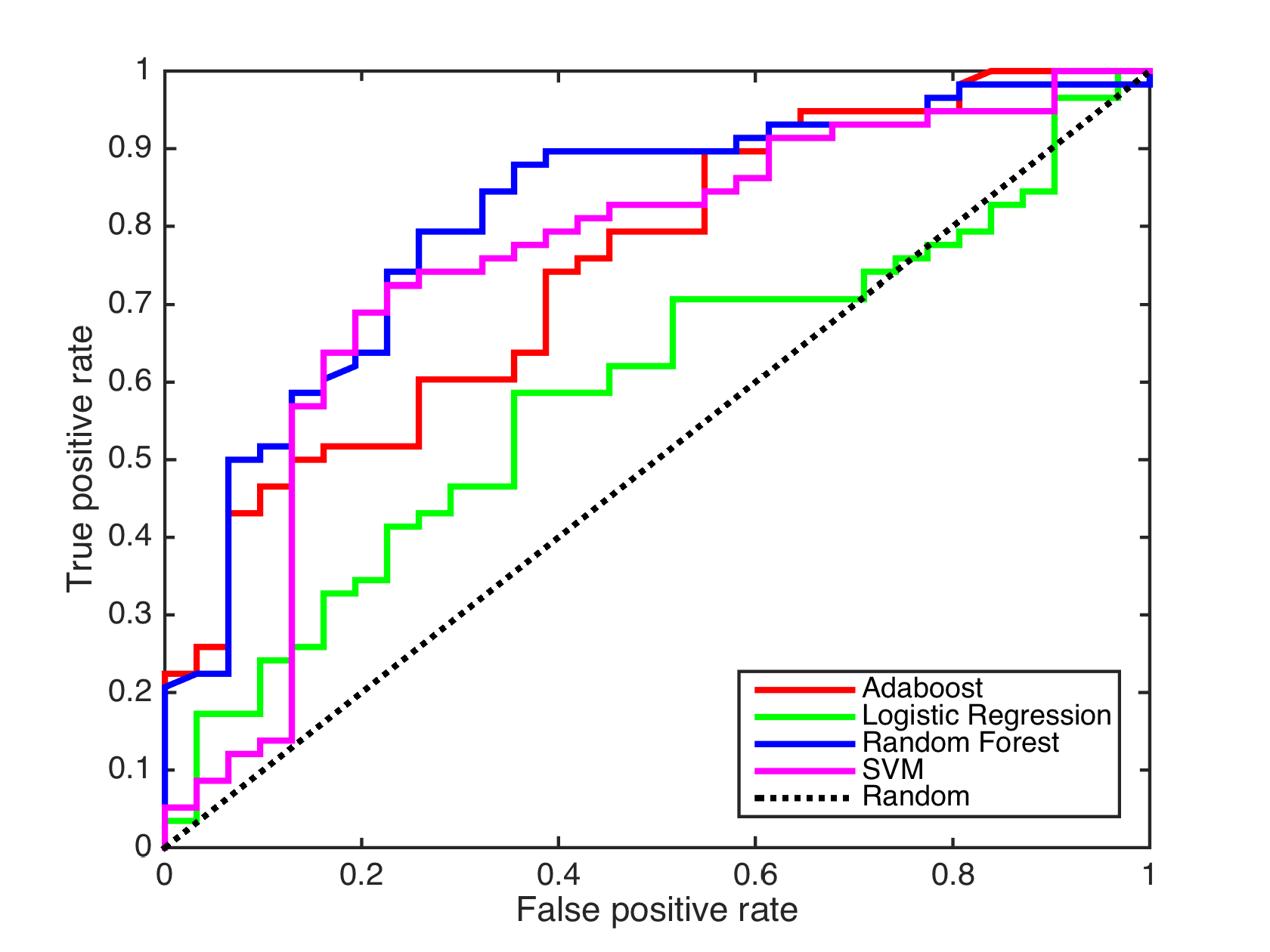}
        \label{fig:roc_SEAL}
    \end{subfigure}
    \label{fig:roc_curve}
\end{figure*}

\begin{table*}
\centering
\caption{Sleep efficiency results}
	\begin{tabular}{c|cc|cc|cc|cc|cc}
	    \hline
	    &\multicolumn{2}{c}{AU-ROC} & \multicolumn{2}{c}{F1 Score}& \multicolumn{2}{c}{Recall}& \multicolumn{2}{c}{Precision}& \multicolumn{2}{c}{Accuracy}
	    \\
	    \hline
	    &				SE+AL	 &	 RAHAR	 & 	SE+AL 	& 	RAHAR 	&	 SE+AL 	& 	RAHAR 	& 	SE+AL 	& 	RAHAR 	&	 SE+AL 	& 	RAHAR 
	    \\
	    \hline
	    \hline
	    Ada		&	0.7489	&	0.8132	&	0.5574	&	0.6885	&	0.5484	&	0.5526	&	0.5667	&	0.9130	&	0.6966	&	0.7206\\
	    RF 		&	0.8115	&	0.8746	&	0.6885	&	0.7500	&	0.6774	&	0.6316	&	0.7000	&	0.9231	&	0.7865	&	0.7647\\
	    SVM		&	0.7497	&	0.7895	&	0.3721	&	0.7077	&	0.2581	&	0.6053	&	0.6667	&	0.8519	&	0.6966	&	0.7206\\
	    LogR 		&	0.5884	&	0.8649	&	-		&	0.6875	&	-		&	0.5789	&	-		&	0.8462	&	-		&	0.7059\\
	    \hline
	    Average	&	0.7246	&	0.8355	&	0.5393*	&	0.7154*	&	0.4946*	&	0.5965*	&	0.6445*	&	0.8960*	&	0.7266*	&	0.7353*\\
	    \hline
	    \multicolumn{11}{@{}l}{{\scriptsize * logistic regression (LogR) score is not included in averaging.}} \\
	\end{tabular}
	\label{tab:slp_eff_res}
\end{table*}

\subsection{Activity Mode Detection}

%\subsection{Change Point Detection}
%Once the time series was segmented into sleep-wake segments, we identified the distinctive activity modes using the multiple change point detection algorithm, hierarchical divisive estimation.  We tested the change points to a statistical significance level of 0.01 and used a maximum number of random permutations of 99. 

%\subsection{Classification and Labeling}
The actigraph accelerometer data contains post-filtered ``counts" for each of the axes. These counts quantify the frequency and intensity of the user's activity\footnote{http://actigraphcorp.com/wp-content/uploads/2015/06/ActiGraph-White-Paper\_What-is-a-Count\_.pdf}. Using Troiano's cut point scale~\cite{troiano2008physical}, the age of a user, and their accelerometer triaxial count, each epoch is labeled with an intensity level: Sedentary, Light, Moderate, or Vigorous. Since each epoch is 1 minute in length, this provides an unnecessary granularity to an individual's activity levels and is highly subject to noise. We ``smooth" the activity intensity levels over activity modes using change point detection.

Once the time series is segmented into sleep-wake segments, we identify the distinctive activity modes using the multiple change point detection algorithm, hierarchical divisive estimation~\cite{NJames:JSS15}. We tested the change points to a statistical significance level of 0.01 and used a maximum number of random permutations of 99. Each change point result is treated as the interval boundaries for distinctive activity modes. 
 
Each sleep-wake segment now consists of a series of change point intervals. The activity intensity classification label for each change point interval is computed by taking the statistical mode of the minute-by-minute labels over every epoch existing within the interval. Fig.~\ref{fig:change_point_labeling} illustrates the classification labelling of an individual's awake time.

\subsection{Modeling}
In sleep science, sleep quality is defined by a number of metrics, including total sleep time, wake after sleep onset, awakening index, and sleep efficiency~\cite{SScholle:SleepMedicine11}. In our analysis, we focus on sleep efficiency metric for our experiments~\cite{JCacioppo:PsycSci02}. Sleep efficiency is computed as a numerical value ranging from 0 to 1. According to specialists, a sleep efficiency below 0.85 (i.e., 85\%) indicates poor sleep quality. Thus, each sleep period can be classified as having ``good sleep efficiency" or ``poor sleep efficiency"~\cite{williams1974electroencephalography}. 

To model the effect of daily physical activity on sleep, the duration of each intensity level of activity was aggregated over the sleep-wake segment. The percentage of awake time in each mode was used as the model input. 

%General intuition, along with preliminary analysis, gave evidence that people's sleep patterns are distinctly different on weekdays and weekends. For this reason, our model uses only weekday sleep-wake activity.  

\section{Experiments and Results}

\begin{figure*}
    \centering
     \caption{Comparison of the performance of random forest model for each approach}
    \begin{subfigure}[b]{0.49\linewidth}        %% or \columnwidth
        \centering
        \caption{ROC}
        \includegraphics[width=\linewidth]{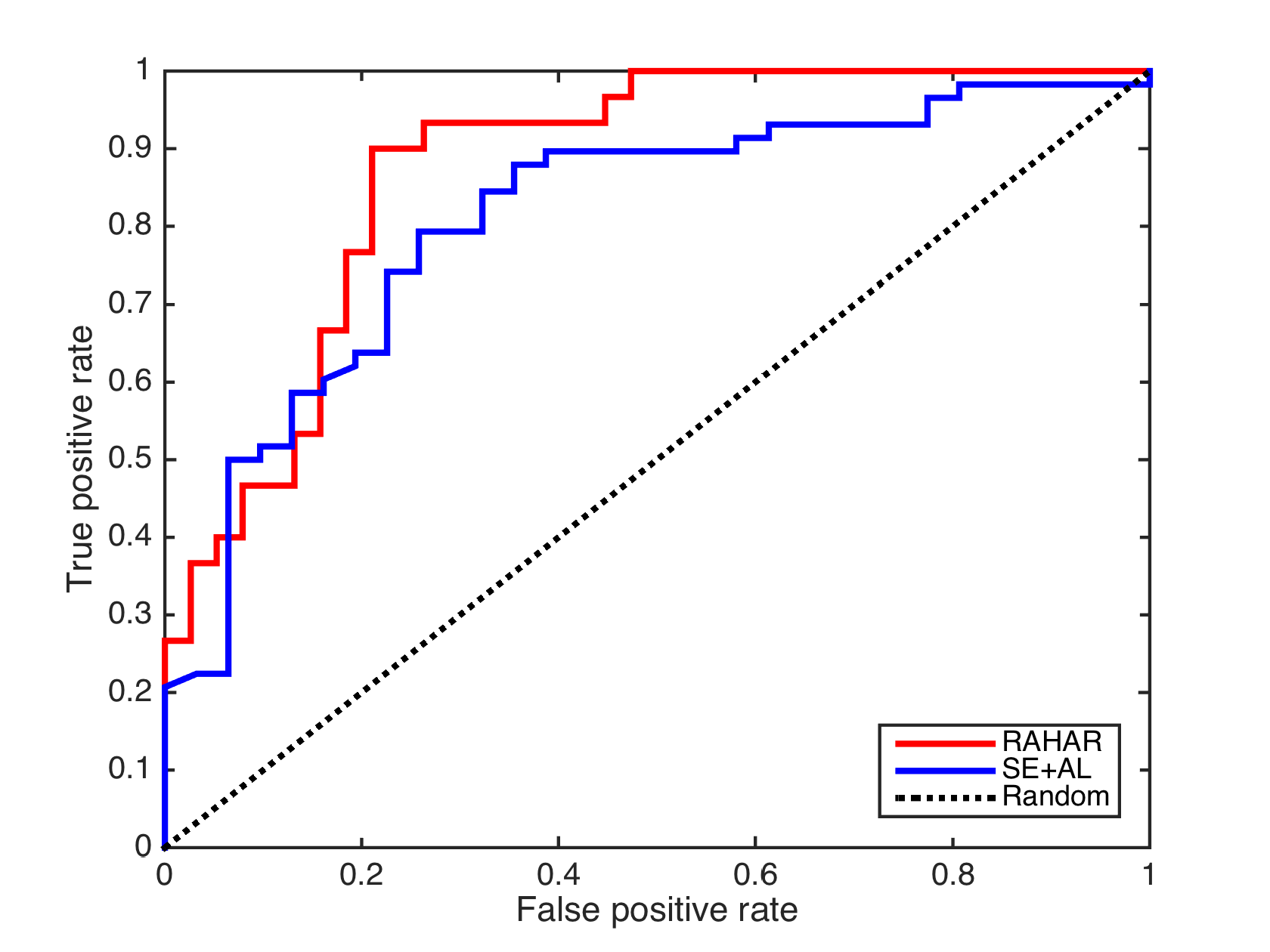}
        \label{fig:rf_roc_comp}
    \end{subfigure}
    \begin{subfigure}[b]{0.49\linewidth}        %% or \columnwidth
        \centering
        \caption{Sensitivity-Specificity}
        \includegraphics[width=\linewidth]{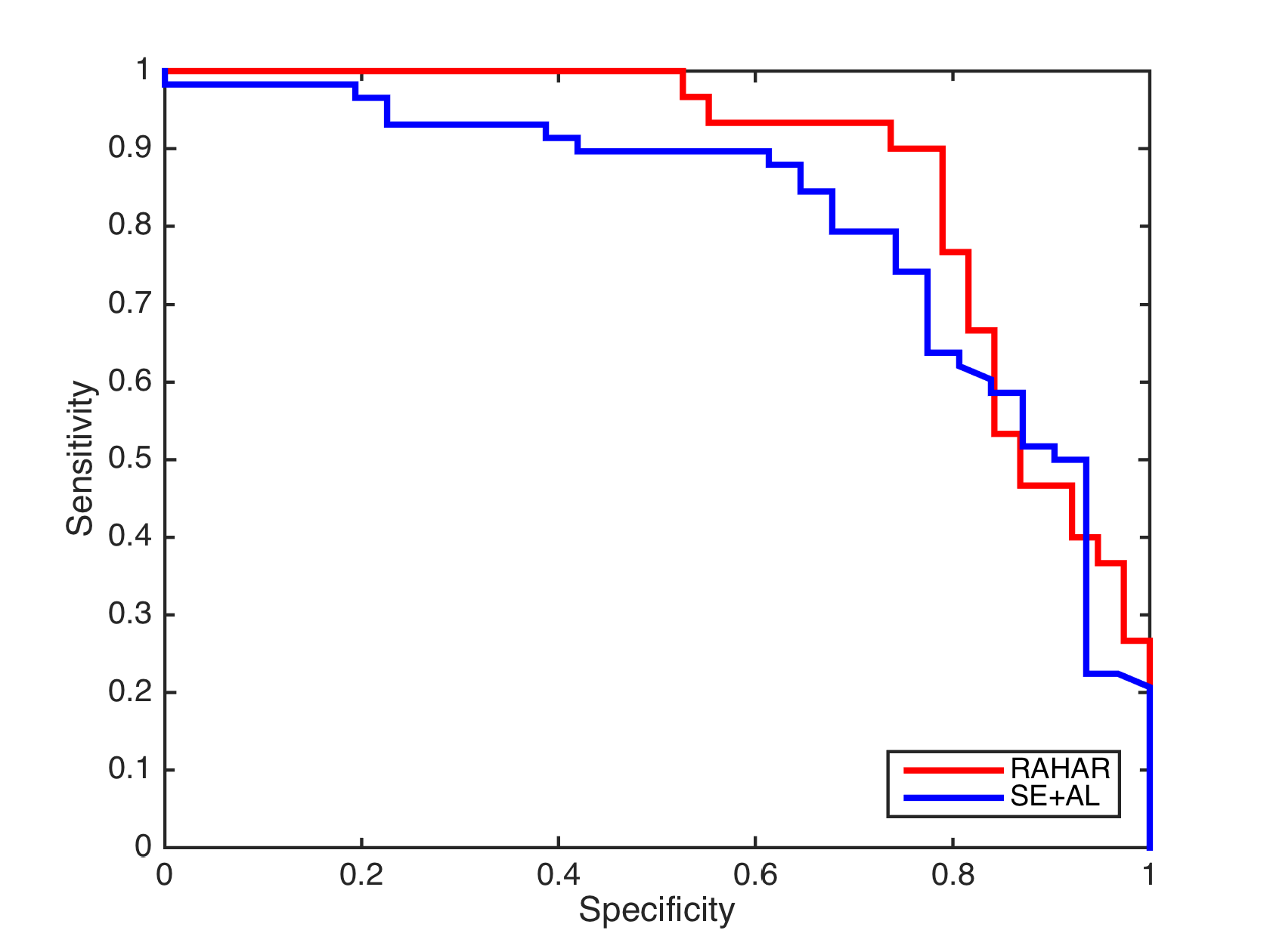}
        \label{fig:rf_ss_comp}
    \end{subfigure}
    \label{fig:rf_comp}
\end{figure*}

\begin{table*}
\centering
\caption{Sleep efficiency -- sensitivity and specificity}
	\begin{tabular}{c|cc|cc|cc|cc}
	    \hline
	    &\multicolumn{2}{c}{AU-ROC} & \multicolumn{2}{c}{F1 Score}& \multicolumn{2}{c}{Sensitivity}& \multicolumn{2}{c}{Specificity}
	    \\
	    \hline
	    &				SE+AL	 &	 RAHAR	 & 	SE+AL 	& 	RAHAR 	&	 SE+AL 	& 	RAHAR 	& 	SE+AL 	& 	RAHAR 
	    \\
	    \hline
	    \hline
	    Ada		&	0.7489	&	0.8132	&	0.5574	&	0.6885	&	0.5484	&	0.5526	&	0.7759	&	0.9333\\
	    RF 		&	0.8115	&	0.8746	&	0.6885	&	0.7500	&	0.6774	&	0.6316	&	0.8448	&	0.9333\\
	    SVM		&	0.7497	&	0.7895	&	0.3721	&	0.7077	&	0.2581	&	0.6053	&	0.9310	&	0.8667\\
	    LogR 		&	0.5884	&	0.8649	&	-		&	0.6875	&	-		&	0.5789	&	-		&	0.8667\\
	    \hline
	    Average	&	0.7246	&	0.8355	&	0.5393*	&	0.7154*	&	0.4946*	&	0.5965*	&	0.8505*	&	0.9111*\\
	    \hline
	    \multicolumn{9}{@{}l}{{\scriptsize * logistic regression (LogR) score is not included in averaging.}} \\
	\end{tabular}
	\label{tab:sens_spec}
\end{table*}

RAHAR is fundamentally a feature extraction algorithm for HAR in the context of quantifying daily physical activity levels of individuals. We therefore test the quality of activity recognition by RAHAR as compared to an expert-based HAR using a tool on continuous physical activity data from a wearable sensor. Since there is \textit{no} ground truth on human activity in this context, our objective is to evaluate which HAR approach leads to better quality models for sleep research, i.e., models for predicting sleep quality, specifically, sleep efficiency.

We selected four models for evaluating the performance of RAHAR against the performance of an expert-based HAR using a tool on the described actigraphy dataset: logistic regression, support vector machines with radial basis function kernel, random forest, and adaboost.

\begin{itemize}
  \item Logistic Regression (LogR): We chose this model because it is an easily interpretable binary classifier. It is also relatively robust to noise, which as explained earlier is a complication on data collected in an uncontrolled environment.\footnote{Even though we included logistic regression (LogR) in our experiments, it is important to note that LogR model failed to stratify the dataset successfully for the state-of-the-art baseline approach, and predicted all cases to be in a single class. Therefore, we excluded LogR score of RAHAR from analysis whenever corresponding LogR score of the state-of-the-art baseline approach was not available.}
   \item Support Vector Machine (SVM): This model was selected because it, also, is a binary classifier. We chose a radial basis function kernel, and so it differs from logistic regression in that it does not linearly divide the data. 
    \item Random Forest (RF): This model was tested as an alternative because of its easy straightforward interpretation, which is particularly relevant in the healthcare or consumer domains. It also is not restricted to linearly dividing the data. 
  \item Adaboost (Ada): Lastly, Adaboost was tested because it is less prone to overfitting than random forest. 
\end{itemize}

For comparison purposes, we use the results from a sleep specialist using Actigraph's ActiLife software as a baseline. The sleep specialist segmentation of the ActiLife results uses the preceding day's activity for each sleep period, and aggregates the activity to an epoch of an hour. ActiLife requires the sleep specialist to manually adjust the sleep period boundaries, and then automatically computes the efficiency and other sleep metrics. 

%Figs.~\ref{fig:roc_RAHAR} and~\ref{fig:roc_SEAL} show the ROC curves for RAHAR and the sleep expert + ActiLife approach, respectively. RAHAR performed best with random forest.

%Table~\ref{tab:slp_eff_res} summarises the results, with ``SE+AL" indicating the sleep expert using the ActiLife software, and RAHAR indicating our automated methodology. As seen in Table~\ref{tab:slp_eff_res}, the sleep expert and ActiLife software have an average AUC  of 0.7700, with the exception of logistic regression. The logistic regression model has an AUC of 0.5884. However, it was unable to stratify the dataset, and so predicted all cases to be in a single class. We consider this to be a failure of logistic regression for this problem, and thus do not consider its results. Our results show, on average, a 15\% improvement from the sleep expert using ActiLife. 
 
Figs.~\ref{fig:roc_RAHAR} and~\ref{fig:roc_SEAL} show the ROC curves for RAHAR and the sleep expert using ActiLife software (denoted as ``SE+AL"), respectively, while Table~\ref{tab:slp_eff_res} summarises the results for both RAHAR and SE+AL. One of the most important performance measures for HAR is the area under ROC (AU-ROC). Based on AU-ROC scores, both RAHAR and SE+AL performed best with random forest model. Furthermore, SE+AL achieved an average AU-ROC of 0.7246 whereas RAHAR achieved 0.8355, a 15\% improvement of AU-ROC score on average by our algorithm as opposed to the sleep expert using ActiLife. With an AU-ROC score of 0.5884 for SE+AL approach, the logistic regression model was, however, unable to stratify the dataset, and so predicted all cases to be in a single class. We considered this to be a failure of the logistic regression model for this problem, and thus, did not include its results in our discussion whenever it was appropriate to do so. For this reason, the misleading results have also been removed from Table~\ref{tab:slp_eff_res}.

Another important performance measure for HAR is the F1 score, which is computed as the harmonic mean of precision and recall. According to Table~\ref{tab:slp_eff_res}, RAHAR performed better than SE+AL in terms of precision and recall for all models, and hence, yielded significantly higher F1 scores. Specifically, F1 score for RAHAR, on average, was 0.7154 whereas it was 0.5393 for SE+AL (excluding logistic regression in both cases), yielding a solid margin of about 0.18 points (i.e., more than 30\% improvement). On the other hand, the accuracy scores, on average, were 0.7353 for RAHAR and 0.7266 for SE+AL (again excluding logistic regression), and exhibited a relatively less significant difference still in favor of RAHAR.

%From figures 3 and 4, it is evident that random forests performed the best. 

%\begin{figure}
%  \includegraphics[width=\linewidth]{Ours_ROC.jpg}
%  \caption{ROC curves of each model for Sleep Efficiency using our novel segmentation methodology}
%\end{figure}

%\begin{figure}
%  \includegraphics[width=\linewidth]{SleepExpert_ROC.jpg}
%  \caption{ROC curves of each model for Sleep Efficiency using current methodology}
%\end{figure}

%\begin{table*}
%\centering
%\caption{Sleep Efficiency Results}
%	\begin{tabular}{c|cc|cc|cc|cc|cc|cc}
%	    \hline
%	    &\multicolumn{2}{c}{ROC} & \multicolumn{2}{c}{F1 Score}& \multicolumn{2}{c}{Sensitivity}& \multicolumn{2}{c}{Specificity}& \multicolumn{2}{c}{Precision}& \multicolumn{2}{c}{Accuracy}
%	    \\
%	    \hline
%	    &				SE+AL	 &	 RAHAR	 & 	SE+AL 	& 	RAHAR 	&	 SE+AL 	& 	RAHAR 	& 	SE+AL 	& 	RAHAR 	&	 SE+AL 	& 	RAHAR	 & 	SE+AL 	& 	RAHAR 
%	    \\
%	    \hline
%	    \hline
%	    Ada		&	0.7489	&	0.8132	&	0.5574	&	0..6885	&	0.5484	&	0..5526	&	0.7759	&	0.9333	&	0.5667	&	0.9130	&	0.6966	&	0.7206\\
%	    RF 		&	0.8115	&	0.8746	&	0.6885	&	0.7302	&	0.6774	&	0.6053	&	0.8448	&	0.9333	&	0.7000	&	0.9200	&	0.7865	&	0.7500\\
%	    SVM		&	0.7497	&	0.7895	&	0.3721	&	0.7077	&	0.2581	&	0.6053	&	0.9310	&	0.7895	&	0.6667	&	0.8519	&	0.6966	&	0.6667\\
%	    Log 		&	0.5884	&	0.8649	&	-		&	0.6875	&	-		&	0.5789	&	-		&	0.8667	&	-		&	0.8462	&	-		&	0.7059\\
%	    \hline
%	\end{tabular}
%\end{table*}

\section{Discussion of Results in Medical Context}

In this section we discuss the results of the best performing model and its broader impact to the area of sleep science. As seen in Fig.~\ref{fig:roc_curve} random forest and logistic regression were the two best performing models with the RAHAR algorithm. Based on the desired threshold value of true positive rate, TPR, (i.e., sensitivity), either model could be preferred to minimize false positive rate, FPR, (i.e., 1-specificity), which is equivalent to maximizing specificity. Random forest was also the best performing model for the SE+AL approach as mentioned earlier. If we compare the ROC as well as the sensitivity-specificity plots of the best model of each approach (i.e., random forest), we see that RAHAR outperforms SE+AL almost always as illustrated in Fig.~\ref{fig:rf_comp}.

Table~\ref{tab:sens_spec}, on the other hand, summarises sensitivity and specificity scores for RAHAR and SE+AL. Average sensitivity score for SE+AL and RAHAR across all models except logistic regression were 0.4946 and 0.5965, respectively. In other words, average sensitivity score for RAHAR is 20\% higher than that of SE+AL. As for specificity, RAHAR with an average score of 0.9111 outperforms SE+AL with an average score of 0.8505, which corresponds to a 7\% improvement.

As we seek to determine in our study whether a person had a ``good quality sleep" based on his physical activity levels during awake period prior to sleep, a false positive occurs when the model predicts ``good quality sleep" when the person actually had a ``poor quality sleep." Therefore, the number of false positives needs to be kept at a minimum for a desired number of true positives. In other words, a high specificity score is sought after while keeping the sensitivity score at the desired level. As can be seen from Fig.~\ref{fig:rf_ss_comp} with this perspective in mind, for a large range of sensitivity scores, RAHAR achieved higher specificity scores almost all the time than SE+AL did. For example, RAHAR achieved a sensitivity score of 0.9 with a specificity score of 0.8 whereas SE+AL remained at a specificity score of 0.6 for the same sensitivity threshold.

In summary, RAHAR outperforms state-of-the-art procedure in sleep research in many aspects. However, its application is not limited to sleep and it can be used for understanding and treatment of other health issues such as obesity, diabetes, or cardiac diseases. Moreover, RAHAR allows for fully automated interpretation without the necessity of manual input or subjective self-reporting.

Given the current interest in deep learning, a natural question that may arise is why an approach based on feature extraction and model building has been used instead of using deep learning models directly on the raw sensor data for HAR. In medical community, the explainability of a model is of utmost important as the medical professionals are interested in learning cause-and-effect relationships and using this knowledge to support their decision making processes. In this particular case, for example, sleep experts are interested in understanding how and when certain physical activity levels effect sleep in order to make decisions to improve sleep quality of individuals accordingly. However, it is an interesting idea to explore deep learning to see what is the best model from a model accuracy perspective to understand the limits of the value of continuous monitoring of individuals' physical activity, not only from a medical perspective in particular but also from a ``quantified self" perspective in general.

%\begin{figure*}
%\centering
%  \caption{ROC curve of Random Forest for each approach}
%  \includegraphics[scale=1]{RandomForest_comp.png}
%  \label{fig:roc_rahar}
%\end{figure*}

\section{Conclusion}
In this paper, we presented a robust automated human activity recognition (RAHAR) algorithm for multi-modal phenomena, and evaluated its performance in the application area of sleep science. We tested the results of RAHAR  against the results of a sleep expert using ActiLife for predicting sleep quality, specifically, sleep efficiency. Our model a) automated the activity recognition, and b) improved the current state-of-the-art results, on average, by ~15\% in terms of AU-ROC and ~30\% in terms of F1 scores across different models. Automating the human activity recognition puts sleep science evaluation in the hands of wearable device users. This empowers users to self-monitor their sleep-wake habits, and take action to improve the quality of their life. The improved results demonstrate the robustness of RAHAR as well as the capabilities of implementing the algorithm within clinical software such as ActiLife. 

The application of RAHAR is, however, not limited to sleep science. It can be used to monitor physical activity levels and patterns of individuals with other health issues such as obesity, diabetes, and cardiac diseases. Besides, RAHAR can also be used in the general context of the ``quantified self" movement, and provide individuals actionable information about their overall fitness levels.

%Moreover, RAHAR allows for fully automated interpretation without the necessity of manual input or subjective self-reporting.

%Future work: fuzzy segmentation, application of HAR to other domains

%\section*{Acknowledgment}

%The authors would like to thank XXX for running the clinical trial wherein the data was collected. 

\bibliographystyle{IEEEtran}
\bibliography{icdm16}

\begin{thebibliography}{10}
\providecommand{\url}[1]{#1}
\csname url@samestyle\endcsname
\providecommand{\newblock}{\relax}
\providecommand{\bibinfo}[2]{#2}
\providecommand{\BIBentrySTDinterwordspacing}{\spaceskip=0pt\relax}
\providecommand{\BIBentryALTinterwordstretchfactor}{4}
\providecommand{\BIBentryALTinterwordspacing}{\spaceskip=\fontdimen2\font plus
\BIBentryALTinterwordstretchfactor\fontdimen3\font minus
  \fontdimen4\font\relax}
\providecommand{\BIBforeignlanguage}[2]{{%
\expandafter\ifx\csname l@#1\endcsname\relax
\typeout{** WARNING: IEEEtran.bst: No hyphenation pattern has been}%
\typeout{** loaded for the language `#1'. Using the pattern for}%
\typeout{** the default language instead.}%
\else
\language=\csname l@#1\endcsname
\fi
#2}}
\providecommand{\BIBdecl}{\relax}
\BIBdecl

\bibitem{HWang:IJCV15}
\BIBentryALTinterwordspacing
H.~Wang, D.~Oneata, J.~Verbeek, and C.~Schmid, ``A robust and efficient video
  representation for action recognition,'' \emph{International Journal of
  Computer Vision}, pp. 1--20, 2015. [Online]. Available:
  \url{http://dx.doi.org/10.1007/s11263-015-0846-5}
\BIBentrySTDinterwordspacing

\bibitem{BFernando:TPAMI16}
B.~Fernando, E.~Gavves, J.~Oramas, A.~Ghodrati, and T.~Tuytelaars, ``Rank
  pooling for action recognition,'' \emph{IEEE Transactions on Pattern Analysis
  and Machine Intelligence}, vol.~PP, no.~99, pp. 1--1, 2016.

\bibitem{IAtmosukarto:WACV15}
I.~Atmosukarto, N.~Ahuja, and B.~Ghanem, ``Action recognition using
  discriminative structured trajectory groups,'' in \emph{2015 IEEE Winter
  Conference on Applications of Computer Vision}, Jan 2015, pp. 899--906.

\bibitem{SMa:CVPR15}
S.~Ma, L.~Sigal, and S.~Sclaroff, ``Space-time tree ensemble for action
  recognition,'' in \emph{The IEEE Conference on Computer Vision and Pattern
  Recognition (CVPR)}, June 2015.

\bibitem{ZZhou:TMM15}
Z.~Zhou, F.~Shi, and W.~Wu, ``Learning spatial and temporal extents of human
  actions for action detection,'' \emph{IEEE Transactions on Multimedia},
  vol.~17, no.~4, pp. 512--525, April 2015.

\bibitem{DTran:ICCV15}
D.~Tran, L.~Bourdev, R.~Fergus, L.~Torresani, and M.~Paluri, ``Learning
  spatiotemporal features with 3d convolutional networks,'' in \emph{2015 IEEE
  International Conference on Computer Vision (ICCV)}, Dec 2015, pp.
  4489--4497.

\bibitem{VKantorov:CVPR14}
V.~Kantorov and I.~Laptev, ``Efficient feature extraction, encoding and
  classification for action recognition,'' in \emph{The IEEE Conference on
  Computer Vision and Pattern Recognition (CVPR)}, June 2014.

\bibitem{XPeng:ECCV14}
\BIBentryALTinterwordspacing
X.~Peng, C.~Zou, Y.~Qiao, and Q.~Peng, \emph{Action Recognition with Stacked
  Fisher Vectors}.\hskip 1em plus 0.5em minus 0.4em\relax Cham: Springer
  International Publishing, 2014, pp. 581--595. [Online]. Available:
  \url{http://dx.doi.org/10.1007/978-3-319-10602-1\_38}
\BIBentrySTDinterwordspacing

\bibitem{HKuehne:WACV16}
H.~Kuehne, J.~Gall, and T.~Serre, ``An end-to-end generative framework for
  video segmentation and recognition,'' in \emph{2016 IEEE Winter Conference on
  Applications of Computer Vision (WACV)}, March 2016, pp. 1--8.

\bibitem{JDonahue:CVPR15}
J.~Donahue, L.~Anne~Hendricks, S.~Guadarrama, M.~Rohrbach, S.~Venugopalan,
  K.~Saenko, and T.~Darrell, ``Long-term recurrent convolutional networks for
  visual recognition and description,'' in \emph{The IEEE Conference on
  Computer Vision and Pattern Recognition (CVPR)}, June 2015.

\bibitem{LWang:CVPR15}
L.~Wang, Y.~Qiao, and X.~Tang, ``Action recognition with trajectory-pooled
  deep-convolutional descriptors,'' in \emph{2015 IEEE Conference on Computer
  Vision and Pattern Recognition (CVPR)}, June 2015, pp. 4305--4314.

\bibitem{LSun:ICCV15}
L.~Sun, K.~Jia, D.~Y. Yeung, and B.~E. Shi, ``Human action recognition using
  factorized spatio-temporal convolutional networks,'' in \emph{2015 IEEE
  International Conference on Computer Vision (ICCV)}, Dec 2015, pp.
  4597--4605.

\bibitem{SKe:Computers13}
S.-R. Ke, H.~L.~U. Thuc, Y.-J. Lee, J.-N. Hwang, J.-H. Yoo, and K.-H. Choi, ``A
  review on video-based human activity recognition,'' \emph{Computers}, vol.~2,
  no.~2, pp. 88--131, 2013.

\bibitem{PBorges:TCSVT13}
P.~V.~K. Borges, N.~Conci, and A.~Cavallaro, ``Video-based human behavior
  understanding: A survey,'' \emph{IEEE Transactions on Circuits and Systems
  for Video Technology}, vol.~23, no.~11, pp. 1993--2008, Nov 2013.

\bibitem{NIkizler:ICPR08}
N.~Ikizler, R.~G. Cinbis, S.~Pehlivan, and P.~Duygulu, ``Recognizing actions
  from still images,'' in \emph{Pattern Recognition, 2008. ICPR 2008. 19th
  International Conference on}, Dec 2008, pp. 1--4.

\bibitem{CThurau:CVPR08}
C.~Thurau and V.~Hlavac, ``Pose primitive based human action recognition in
  videos or still images,'' in \emph{Computer Vision and Pattern Recognition,
  2008. CVPR 2008. IEEE Conference on}, June 2008, pp. 1--8.

\bibitem{FKhan:IJCV13}
\BIBentryALTinterwordspacing
F.~S. Khan, R.~Muhammad~Anwer, J.~van~de Weijer, A.~D. Bagdanov, A.~M. Lopez,
  and M.~Felsberg, ``Coloring action recognition in still images,''
  \emph{International Journal of Computer Vision}, vol. 105, no.~3, pp.
  205--221, 2013. [Online]. Available:
  \url{http://dx.doi.org/10.1007/s11263-013-0633-0}
\BIBentrySTDinterwordspacing

\bibitem{WYang:CVPR10}
W.~Yang, Y.~Wang, and G.~Mori, ``Recognizing human actions from still images
  with latent poses,'' in \emph{Computer Vision and Pattern Recognition (CVPR),
  2010 IEEE Conference on}, June 2010, pp. 2030--2037.

\bibitem{SMaji:CVPR11}
S.~Maji, L.~Bourdev, and J.~Malik, ``Action recognition from a distributed
  representation of pose and appearance,'' in \emph{Computer Vision and Pattern
  Recognition (CVPR), 2011 IEEE Conference on}, June 2011, pp. 3177--3184.

\bibitem{BYao:ICCV11}
B.~Yao, X.~Jiang, A.~Khosla, A.~L. Lin, L.~Guibas, and L.~Fei-Fei, ``Human
  action recognition by learning bases of action attributes and parts,'' in
  \emph{2011 International Conference on Computer Vision}, Nov 2011, pp.
  1331--1338.

\bibitem{GSharma:TPAMI16}
G.~Sharma, F.~Jurie, and C.~Schmid, ``Expanded parts model for semantic
  description of humans in still images,'' \emph{IEEE Transactions on Pattern
  Analysis and Machine Intelligence}, vol.~PP, no.~99, pp. 1--1, 2016.

\bibitem{GGuo:PatRec14}
\BIBentryALTinterwordspacing
G.~Guo and A.~Lai, ``A survey on still image based human action recognition,''
  \emph{Pattern Recognition}, vol.~47, no.~10, pp. 3343 -- 3361, 2014.
  [Online]. Available:
  \url{http://www.sciencedirect.com/science/article/pii/S0031320314001642}
\BIBentrySTDinterwordspacing

\bibitem{WLi:CVPRW10}
W.~Li, Z.~Zhang, and Z.~Liu, ``Action recognition based on a bag of 3d
  points,'' in \emph{2010 IEEE Computer Society Conference on Computer Vision
  and Pattern Recognition - Workshops}, June 2010, pp. 9--14.

\bibitem{LXia:CVPRW12}
L.~Xia, C.~C. Chen, and J.~K. Aggarwal, ``View invariant human action
  recognition using histograms of 3d joints,'' in \emph{2012 IEEE Computer
  Society Conference on Computer Vision and Pattern Recognition Workshops},
  June 2012, pp. 20--27.

\bibitem{XYang:CVPRW12}
X.~Yang and Y.~L. Tian, ``Eigenjoints-based action recognition using
  naive-bayes-nearest-neighbor,'' in \emph{2012 IEEE Computer Society
  Conference on Computer Vision and Pattern Recognition Workshops}, June 2012,
  pp. 14--19.

\bibitem{FOfli:JVCI14}
\BIBentryALTinterwordspacing
F.~Ofli, R.~Chaudhry, G.~Kurillo, R.~Vidal, and R.~Bajcsy, ``Sequence of the
  most informative joints (smij): A new representation for human skeletal
  action recognition,'' \emph{Journal of Visual Communication and Image
  Representation}, vol.~25, no.~1, pp. 24 -- 38, 2014, visual Understanding and
  Applications with RGB-D Cameras. [Online]. Available:
  \url{http://www.sciencedirect.com/science/article/pii/S1047320313000680}
\BIBentrySTDinterwordspacing

\bibitem{RChaudhry:CVPRW13}
R.~Chaudhry, F.~Ofli, G.~Kurillo, R.~Bajcsy, and R.~Vidal, ``Bio-inspired
  dynamic 3d discriminative skeletal features for human action recognition,''
  in \emph{2013 IEEE Conference on Computer Vision and Pattern Recognition
  Workshops}, June 2013, pp. 471--478.

\bibitem{JWang:CVPR12}
J.~Wang, Z.~Liu, Y.~Wu, and J.~Yuan, ``Mining actionlet ensemble for action
  recognition with depth cameras,'' in \emph{Computer Vision and Pattern
  Recognition (CVPR), 2012 IEEE Conference on}, June 2012, pp. 1290--1297.

\bibitem{RVemulapalli:CVPR14}
R.~Vemulapalli, F.~Arrate, and R.~Chellappa, ``Human action recognition by
  representing 3d skeletons as points in a lie group,'' in \emph{2014 IEEE
  Conference on Computer Vision and Pattern Recognition}, June 2014, pp.
  588--595.

\bibitem{MHussein:IJCAI13}
M.~E. Hussein, M.~Torki, M.~A. Gowayyed, and M.~El-Saban, ``Human action
  recognition using a temporal hierarchy of covariance descriptors on 3d joint
  locations,'' \emph{IJCAI}, vol.~13, pp. 2466--2472, 2013.

\bibitem{FLv:ECCV06}
\BIBentryALTinterwordspacing
F.~Lv and R.~Nevatia, \emph{Recognition and Segmentation of 3-D Human Action
  Using HMM and Multi-class AdaBoost}.\hskip 1em plus 0.5em minus 0.4em\relax
  Berlin, Heidelberg: Springer Berlin Heidelberg, 2006, pp. 359--372. [Online].
  Available: \url{http://dx.doi.org/10.1007/11744085\_28}
\BIBentrySTDinterwordspacing

\bibitem{YSheikh:ICCV05}
Y.~Sheikh, M.~Sheikh, and M.~Shah, ``Exploring the space of a human action,''
  in \emph{Tenth IEEE International Conference on Computer Vision (ICCV'05)
  Volume 1}, vol.~1, Oct 2005, pp. 144--149 Vol. 1.

\bibitem{JWang:ECCV12}
\BIBentryALTinterwordspacing
J.~Wang, Z.~Liu, J.~Chorowski, Z.~Chen, and Y.~Wu, \emph{Robust 3D Action
  Recognition with Random Occupancy Patterns}.\hskip 1em plus 0.5em minus
  0.4em\relax Berlin, Heidelberg: Springer Berlin Heidelberg, 2012, pp.
  872--885. [Online]. Available:
  \url{http://dx.doi.org/10.1007/978-3-642-33709-3\_62}
\BIBentrySTDinterwordspacing

\bibitem{AVieira:CIARP12}
\BIBentryALTinterwordspacing
A.~W. Vieira, E.~R. Nascimento, G.~L. Oliveira, Z.~Liu, and M.~F.~M. Campos,
  \emph{STOP: Space-Time Occupancy Patterns for 3D Action Recognition from
  Depth Map Sequences}.\hskip 1em plus 0.5em minus 0.4em\relax Berlin,
  Heidelberg: Springer Berlin Heidelberg, 2012, pp. 252--259. [Online].
  Available: \url{http://dx.doi.org/10.1007/978-3-642-33275-3\_31}
\BIBentrySTDinterwordspacing

\bibitem{JAggarwal:PatRec14}
\BIBentryALTinterwordspacing
J.~Aggarwal and L.~Xia, ``Human activity recognition from 3d data: A review,''
  \emph{Pattern Recognition Letters}, vol.~48, pp. 70 -- 80, 2014, celebrating
  the life and work of Maria Petrou. [Online]. Available:
  \url{http://www.sciencedirect.com/science/article/pii/S0167865514001299}
\BIBentrySTDinterwordspacing

\bibitem{LChen:TCMCC12}
L.~Chen, J.~Hoey, C.~D. Nugent, D.~J. Cook, and Z.~Yu, ``Sensor-based activity
  recognition,'' \emph{IEEE Transactions on Systems, Man, and Cybernetics, Part
  C (Applications and Reviews)}, vol.~42, no.~6, pp. 790--808, Nov 2012.

\bibitem{ABulling:CSUR14}
\BIBentryALTinterwordspacing
A.~Bulling, U.~Blanke, and B.~Schiele, ``A tutorial on human activity
  recognition using body-worn inertial sensors,'' \emph{ACM Comput. Surv.},
  vol.~46, no.~3, pp. 33:1--33:33, Jan. 2014. [Online]. Available:
  \url{http://doi.acm.org/10.1145/2499621}
\BIBentrySTDinterwordspacing

\bibitem{ACalzada:EMBC14}
A.~Calzada, J.~Liu, C.~D. Nugent, H.~Wang, and L.~Martinez, ``Sensor-based
  activity recognition using extended belief rule-based inference
  methodology,'' in \emph{2014 36th Annual International Conference of the IEEE
  Engineering in Medicine and Biology Society}, Aug 2014, pp. 2694--2697.

\bibitem{DBiswas:HUMOV15}
\BIBentryALTinterwordspacing
D.~Biswas, A.~Cranny, N.~Gupta, K.~Maharatna, J.~Achner, J.~Klemke, M.~Jöbges,
  and S.~Ortmann, ``Recognizing upper limb movements with wrist worn inertial
  sensors using k-means clustering classification,'' \emph{Human Movement
  Science}, vol.~40, pp. 59 -- 76, 2015. [Online]. Available:
  \url{http://www.sciencedirect.com/science/article/pii/S0167945714002115}
\BIBentrySTDinterwordspacing

\bibitem{ATripathi:EAIS15}
A.~M. Tripathi, D.~Baruah, and R.~D. Baruah, ``Acoustic sensor based activity
  recognition using ensemble of one-class classifiers,'' in \emph{Evolving and
  Adaptive Intelligent Systems (EAIS), 2015 IEEE International Conference on},
  Dec 2015, pp. 1--7.

\bibitem{CCatal:ASOC15}
\BIBentryALTinterwordspacing
C.~Catal, S.~Tufekci, E.~Pirmit, and G.~Kocabag, ``On the use of ensemble of
  classifiers for accelerometer-based activity recognition,'' \emph{Applied
  Soft Computing}, vol.~37, pp. 1018 -- 1022, 2015. [Online]. Available:
  \url{http://www.sciencedirect.com/science/article/pii/S1568494615000447}
\BIBentrySTDinterwordspacing

\bibitem{RAkhavian:WSC15}
R.~Akhavian and A.~Behzadan, ``Wearable sensor-based activity recognition for
  data-driven simulation of construction workers' activities,'' in \emph{2015
  Winter Simulation Conference (WSC)}, Dec 2015, pp. 3333--3344.

\bibitem{LLiu:KNOSYS15}
\BIBentryALTinterwordspacing
L.~Liu, Y.~Peng, M.~Liu, and Z.~Huang, ``Sensor-based human activity
  recognition system with a multilayered model using time series shapelets,''
  \emph{Knowledge-Based Systems}, vol.~90, pp. 138 -- 152, 2015. [Online].
  Available:
  \url{http://www.sciencedirect.com/science/article/pii/S0950705115003639}
\BIBentrySTDinterwordspacing

\bibitem{GOkeyo:PMCJ14}
\BIBentryALTinterwordspacing
G.~Okeyo, L.~Chen, H.~Wang, and R.~Sterritt, ``Dynamic sensor data segmentation
  for real-time knowledge-driven activity recognition,'' \emph{Pervasive and
  Mobile Computing}, vol. 10, Part B, pp. 155 -- 172, 2014. [Online].
  Available:
  \url{http://www.sciencedirect.com/science/article/pii/S1574119212001393}
\BIBentrySTDinterwordspacing

\bibitem{BDobkin:CurrOpinNeurol13}
B.~H. Dobkin, ``Wearable motion sensors to continuously measure real-world
  physical activities,'' \emph{Current opinion in neurology}, vol.~26, no.~6,
  pp. 602--608, 2013.

\bibitem{OLara:SURV13}
O.~D. Lara and M.~A. Labrador, ``A survey on human activity recognition using
  wearable sensors,'' \emph{IEEE Communications Surveys Tutorials}, vol.~15,
  no.~3, pp. 1192--1209, Third 2013.

\bibitem{JBort-Roig:Sports14}
\BIBentryALTinterwordspacing
J.~Bort-Roig, N.~D. Gilson, A.~Puig-Ribera, R.~S. Contreras, and S.~G. Trost,
  ``Measuring and influencing physical activity with smartphone technology: A
  systematic review,'' \emph{Sports Medicine}, vol.~44, no.~5, pp. 671--686,
  2014. [Online]. Available: \url{http://dx.doi.org/10.1007/s40279-014-0142-5}
\BIBentrySTDinterwordspacing

\bibitem{MRehman:Sensors15}
M.~H. ur~Rehman, C.~S. Liew, T.~Y. Wah, J.~Shuja, and B.~Daghighi, ``Mining
  personal data using smartphones and wearable devices: A survey,''
  \emph{Sensors}, vol.~15, no.~2, pp. 4430--4469, 2015.

\bibitem{SMukhopadhyay:JSEN15}
S.~C. Mukhopadhyay, ``Wearable sensors for human activity monitoring: A
  review,'' \emph{IEEE Sensors Journal}, vol.~15, no.~3, pp. 1321--1330, March
  2015.

\bibitem{MZhou:ICMR13}
\BIBentryALTinterwordspacing
L.~M. Zhou, C.~Gurrin, and Z.~Qiu, ``Zhiwo: Activity tagging and recognition
  system for personal lifelogs,'' in \emph{Proceedings of the 3rd ACM
  Conference on International Conference on Multimedia Retrieval}, ser. ICMR
  '13.\hskip 1em plus 0.5em minus 0.4em\relax New York, NY, USA: ACM, 2013, pp.
  321--322. [Online]. Available:
  \url{http://doi.acm.org/10.1145/2461466.2461526}
\BIBentrySTDinterwordspacing

\bibitem{CMeurisch:UbiComp15}
\BIBentryALTinterwordspacing
C.~Meurisch, B.~Schmidt, M.~Scholz, I.~Schweizer, and M.~M\"{u}hlh\"{a}user,
  ``Labels: Quantified self app for human activity sensing,'' in \emph{Adjunct
  Proceedings of the 2015 ACM International Joint Conference on Pervasive and
  Ubiquitous Computing and Proceedings of the 2015 ACM International Symposium
  on Wearable Computers}, ser. UbiComp/ISWC'15 Adjunct.\hskip 1em plus 0.5em
  minus 0.4em\relax New York, NY, USA: ACM, 2015, pp. 1413--1422. [Online].
  Available: \url{http://doi.acm.org/10.1145/2800835.2801612}
\BIBentrySTDinterwordspacing

\bibitem{JHamm:MobiCASE13}
\BIBentryALTinterwordspacing
J.~Hamm, B.~Stone, M.~Belkin, and S.~Dennis, \emph{Automatic Annotation of
  Daily Activity from Smartphone-Based Multisensory Streams}.\hskip 1em plus
  0.5em minus 0.4em\relax Berlin, Heidelberg: Springer Berlin Heidelberg, 2013,
  pp. 328--342. [Online]. Available:
  \url{http://dx.doi.org/10.1007/978-3-642-36632-1\_19}
\BIBentrySTDinterwordspacing

\bibitem{CDobbins:CIT15}
C.~Dobbins and R.~Rawassizadeh, ``Clustering of physical activities for
  quantified self and mhealth applications,'' in \emph{Computer and Information
  Technology; Ubiquitous Computing and Communications; Dependable, Autonomic
  and Secure Computing; Pervasive Intelligence and Computing
  (CIT/IUCC/DASC/PICOM), 2015 IEEE International Conference on}, Oct 2015, pp.
  1423--1428.

\bibitem{MUddin:WearSys15}
\BIBentryALTinterwordspacing
M.~Uddin, A.~Salem, I.~Nam, and T.~Nadeem, ``Wearable sensing framework for
  human activity monitoring,'' in \emph{Proceedings of the 2015 Workshop on
  Wearable Systems and Applications}, ser. WearSys '15.\hskip 1em plus 0.5em
  minus 0.4em\relax New York, NY, USA: ACM, 2015, pp. 21--26. [Online].
  Available: \url{http://doi.acm.org/10.1145/2753509.2753513}
\BIBentrySTDinterwordspacing

\bibitem{OBanos:EMBC15}
O.~Banos, J.~Bang, T.~Hur, M.~H. Siddiqi, H.~T. Thien, L.~B. Vui, W.~A. Khan,
  T.~Ali, C.~Villalonga, and S.~Lee, ``Mining human behavior for health
  promotion,'' in \emph{2015 37th Annual International Conference of the IEEE
  Engineering in Medicine and Biology Society (EMBC)}, Aug 2015, pp.
  5062--5065.

\bibitem{strine2005associations}
T.~W. Strine and D.~P. Chapman, ``Associations of frequent sleep insufficiency
  with health-related quality of life and health behaviors,'' \emph{Sleep
  medicine}, vol.~6, no.~1, pp. 23--27, 2005.

\bibitem{arora2013investigation}
T.~Arora, E.~Broglia, D.~Pushpakumar, T.~Lodhi, and S.~Taheri, ``An
  investigation into the strength of the association and agreement levels
  between subjective and objective sleep duration in adolescents,'' \emph{PloS
  one}, vol.~8, no.~8, p. e72406, 2013.

\bibitem{PFreedson:MedSciSports98}
P.~S. Freedson, E.~Melanson, and J.~Sirard, ``Calibration of the computer
  science and applications inc. accelerometer,'' \emph{Medicine and Science in
  Sports and Exercise}, vol.~30, no.~5, pp. 777--781, 1998.

\bibitem{CIber:AASM07}
C.~Iber, S.~Ancoli-Israel, A.~Chesson, and S.~Quan, \emph{The {AASM} Manual for
  the Scoring of Sleep and Associated Events: Rules, Terminology and Technical
  Specifications}.\hskip 1em plus 0.5em minus 0.4em\relax American Academy of
  Sleep Medicine, 2007.

\bibitem{sadeh2000sleep}
A.~Sadeh, A.~Raviv, and R.~Gruber, ``Sleep patterns and sleep disruptions in
  school-age children,'' \emph{Developmental psychology}, vol.~36, no.~3, p.
  291, 2000.

\bibitem{JCacioppo:PsycSci02}
\BIBentryALTinterwordspacing
J.~T. Cacioppo, L.~C. Hawkley, G.~G. Berntson, J.~M. Ernst, A.~C. Gibbs,
  R.~Stickgold, and J.~A. Hobson, ``Do lonely days invade the nights? potential
  social modulation of sleep efficiency,'' \emph{Psychological Science},
  vol.~13, no.~4, pp. 384--387, 2002. [Online]. Available:
  \url{http://pss.sagepub.com/content/13/4/384.abstract}
\BIBentrySTDinterwordspacing

\bibitem{SScholle:SleepMedicine11}
\BIBentryALTinterwordspacing
S.~Scholle, U.~Beyer, M.~Bernhard, S.~Eichholz, T.~Erler, P.~Graneß,
  B.~Goldmann-Schnalke, K.~Heisch, F.~Kirchhoff, K.~Klementz, G.~Koch,
  A.~Kramer, C.~Schmidtlein, B.~Schneider, B.~Walther, A.~Wiater, and H.~C.
  Scholle, ``Normative values of polysomnographic parameters in childhood and
  adolescence: Quantitative sleep parameters,'' \emph{Sleep Medicine}, vol.~12,
  no.~6, pp. 542 -- 549, 2011. [Online]. Available:
  \url{http://www.sciencedirect.com/science/article/pii/S1389945711001122}
\BIBentrySTDinterwordspacing

\bibitem{KGabriel:IJBNPA10}
\BIBentryALTinterwordspacing
K.~P. Gabriel, J.~J. McClain, K.~K. Schmid, K.~L. Storti, R.~R. High, D.~A.
  Underwood, L.~H. Kuller, and A.~M. Kriska, ``Issues in accelerometer
  methodology: the role of epoch length on estimates of physical activity and
  relationships with health outcomes in overweight, post-menopausal women,''
  \emph{International Journal of Behavioral Nutrition and Physical Activity},
  vol.~7, no.~1, pp. 1--10, 2010. [Online]. Available:
  \url{http://dx.doi.org/10.1186/1479-5868-7-53}
\BIBentrySTDinterwordspacing

\bibitem{troiano2008physical}
R.~P. Troiano, D.~Berrigan, K.~W. Dodd, L.~C. Masse, T.~Tilert, M.~McDowell
  \emph{et~al.}, ``Physical activity in the united states measured by
  accelerometer,'' \emph{Medicine and science in sports and exercise}, vol.~40,
  no.~1, pp. 181--188, 2008.

\bibitem{NJames:JSS15}
\BIBentryALTinterwordspacing
N.~James and D.~Matteson, ``ecp: An r package for nonparametric multiple change
  point analysis of multivariate data,'' \emph{Journal of Statistical
  Software}, vol.~62, no.~1, pp. 1--25, 2015. [Online]. Available:
  \url{https://www.jstatsoft.org/index.php/jss/article/view/v062i07}
\BIBentrySTDinterwordspacing

\bibitem{williams1974electroencephalography}
\BIBentryALTinterwordspacing
R.~Williams, I.~Karacan, and C.~Hursch, \emph{Electroencephalography (Eeg) of
  Human Sleep: Clinical Applications}, ser. A Wiley biomedical-health
  publication.\hskip 1em plus 0.5em minus 0.4em\relax Wiley, 1974. [Online].
  Available: \url{https://books.google.com.qa/books?id=xv1rAAAAMAAJ}
\BIBentrySTDinterwordspacing

\end{thebibliography}

\end{document}